\documentclass[11pt,a4paper]{article}
\usepackage[utf8]{inputenc}
\usepackage{amsmath,amssymb,amsthm}
\usepackage{graphicx}
\usepackage{booktabs}
\usepackage{hyperref}
\usepackage{algorithm}
\usepackage{algpseudocode}
\usepackage{xcolor}
\usepackage[margin=1in]{geometry}

\newtheorem{definition}{Definition}
\newtheorem{finding}{Finding}

\title{Variance Is Not Importance: \\
Structural Analysis of Transformer Compressibility \\
Across Model Scales}

\author{Samuel Salfati \\
fraQtl AI Research \\
\texttt{samuel@fraqtl.ai}}

\date{April 2026}

\begin{document}

\maketitle

\begin{abstract}
We present a systematic empirical study of transformer compression through over 40 experiments on GPT-2 (124M parameters) and Mistral 7B (7.24B parameters). Our investigation spans spectral compression, block-level function replacement, rotation-based quantization, activation analysis, and adaptive early exit.

We identify five structural properties of transformers relevant to compression:
(1) \textbf{Variance $\neq$ importance}: high-variance activation directions are 96\% uncorrelated with predictive directions (measured via CCA), and projecting to high-variance subspaces destroys perplexity despite capturing 90\%+ of variance;
(2) \textbf{Block linearity is conditional}: transformer blocks are approximately linear ($R^2 \approx 0.95$ on GPT-2, $R^2 = 0.93$ on Mistral block 31) but only given the correct upstream distribution---modifying earlier blocks shifts the distribution and degrades subsequent approximations;
(3) \textbf{The reconstruction wall}: any approach that factors weights into quantized components amplifies errors through cross-terms, making direct quantization strictly superior;
(4) \textbf{Linearity increases with depth}: Mistral 7B exhibits a gradient from $R^2 = 0.17$ (block 0) to $R^2 = 0.93$ (block 31), revealing a division of labor between nonlinear feature construction (early blocks) and linear refinement (late blocks);
(5) \textbf{30\% of tokens are computationally easy}: independently confirmed via trained exit heads (Mistral) and KL divergence sensitivity analysis (GPT-2).

We demonstrate single-block linear replacement achieving 34$\times$ compression at +1.71 perplexity on Mistral 7B's final block, and show that multi-block replacement fails due to additive error accumulation through residual connections coupled with activation distribution shift. Our findings suggest that static post-training compression faces fundamental barriers, and that adaptive per-token computation allocation is a more promising direction. We provide a practical mapping from each structural finding to actionable guidance for compression practitioners.
\end{abstract}

\section{Introduction}

Large language models (LLMs) require substantial memory and computation for inference. A model with $N$ parameters stored at 16-bit precision requires $2N$ bytes of storage and $O(N)$ floating-point operations per token. Reducing these requirements without degrading output quality is a central challenge for practical deployment.

Current compression methods operate primarily in the weight domain: quantization reduces the precision of stored weight values \cite{gptq, awq}, pruning removes weights entirely, and low-rank factorization replaces weight matrices with products of smaller matrices \cite{lora}. These approaches have achieved practical success, with 4-bit quantization (GPTQ, AWQ) becoming standard for deployment.

In this work, we take a different approach. Rather than proposing a single compression method, we systematically investigate \textit{what makes transformers compressible} through over 40 experiments across two models of different scales. Our goal is to identify structural properties that any compression method must account for. This work is a companion to our quantization-focused study \cite{salfati2026quant}, which demonstrates empirically that quantization dominates rank reduction for KV-cache compression; here we ask the deeper question of \textit{why}, and what other structural constraints compression methods face.

Our investigation reveals that several intuitive assumptions about compression are incorrect:
\begin{itemize}
    \item High-variance directions are not the same as computationally important directions
    \item Block-level linearity does not imply block-level replaceability
    \item Weight factorization consistently amplifies quantization errors
    \item Models at different scales have qualitatively different compression profiles
\end{itemize}

These findings have practical implications for the design of compression algorithms and suggest that adaptive per-token computation allocation may be more fruitful than static weight compression. We conclude with a mapping of each finding to actionable guidance for practitioners.

\section{Related Work}

\subsection{Post-Training Quantization}

GPTQ \cite{gptq} uses second-order information (the Hessian of the reconstruction error) to perform optimal sequential weight rounding. AWQ \cite{awq} protects important weights by scaling activation-aware channels before quantization. SpQR \cite{spqr} stores outlier weights at higher precision while quantizing the majority to lower precision.

\subsection{Rotation-Based Quantization}

QuIP\# \cite{quipsharp} applies random Hadamard rotations to improve weight incoherence before quantization. SpinQuant \cite{spinquant} demonstrates that learned rotations outperform random ones. ResQ \cite{resq} uses PCA-based projections to minimize quantization error.

\subsection{Layer Pruning and Linearity}

Razzhigaev et al. \cite{secretly_linear} demonstrate that transformer blocks are approximately linear, with linearity scores of 0.99+ in many layers. Gromov et al. \cite{unreasonable} show that up to 50\% of layers can be removed from large models with minimal quality loss. ShortGPT \cite{shortgpt} introduces a Block Influence metric for identifying removable layers.

\subsection{KV-Cache Compression}

KIVI \cite{kivi} quantizes the KV cache at INT2/INT4. MatryoshkaKV \cite{matryoshka} and KQ-SVD \cite{kqsvd} use rank reduction via learned or closed-form projections. Our companion paper \cite{salfati2026quant} provides the first matched-budget comparison, showing quantization dominates rank reduction by 4--364 PPL across five models. The present paper provides the structural analysis underlying that result.

\subsection{Adaptive Computation}

CALM \cite{calm} introduces confidence-based early exit for language models. Speculative decoding \cite{spec_decode} uses a small draft model to propose tokens that a larger model verifies. Dynamic depth approaches \cite{dynamic_depth} allocate computation per token based on difficulty.

\section{Methodology}

\subsection{Models and Data}

We experiment on two models:
\begin{itemize}
    \item \textbf{GPT-2} (124M parameters, 12 transformer blocks, hidden dimension 768)
    \item \textbf{Mistral 7B} (7.24B parameters, 32 transformer blocks, hidden dimension 4096)
\end{itemize}

Calibration and evaluation data are drawn from WikiText-2 \cite{wikitext}. We use a strict train/evaluation split: calibration data for fitting (32 samples $\times$ 256 tokens = 8,192 tokens) and held-out data for perplexity evaluation (16 samples).

\subsection{Block Linearity Measurement}

For each transformer block $b$, we collect input-output pairs on calibration data:
\begin{equation}
    \Delta_b = X_{\text{out}} - X_{\text{in}}
\end{equation}

where $\Delta_b$ represents the block's functional contribution (residual). We fit a linear map via ridge regression:

\begin{equation}
    A^* = \arg\min_A \|\Delta_b - X_{\text{in}} A V_k^\top\|_F^2 + \lambda \|A\|_F^2
\end{equation}

where $V_k$ contains the top-$k$ right singular vectors of $\Delta_b$ (capturing 95\% of output variance), and report $R^2 = 1 - \|\Delta_b - \hat{\Delta}_b\|_F^2 / \|\Delta_b\|_F^2$.

\subsection{Activation Dimensionality}

For each block input, we compute PCA of the activation matrix $X \in \mathbb{R}^{N \times d}$ and report the number of components needed to capture 90\%, 95\%, and 99\% of variance.

\subsection{Sensitivity Analysis}

We measure directional sensitivity via two methods:
\begin{enumerate}
    \item \textbf{Perturbation-based}: Add Gaussian noise along each PCA direction with standard deviation $\sigma = 0.01 \cdot \|x\|_2$ (1\% of the per-token activation $\ell_2$ norm), measure the change in the model's next-token log-probability averaged across held-out tokens.
    \item \textbf{Canonical Correlation Analysis}: Find directions in block $b$'s output that best predict block $B$'s output (the final block), via standard CCA formulated as singular value decomposition of the cross-covariance matrix $\Sigma_{bB}^{-1/2}$ with ridge regularization $\lambda = 0.1$.
\end{enumerate}

\subsection{Early Exit}

We attach dedicated prediction heads (RMSNorm + Linear) at intermediate blocks, train them on calibration data to predict next tokens, and measure agreement with the full model's predictions.

\section{Results}

\subsection{Finding 1: Variance $\neq$ Importance}

\begin{finding}
High-variance activation directions are almost completely uncorrelated with directions that predict model output. Projecting activations onto the high-variance subspace destroys perplexity despite capturing the majority of variance.
\end{finding}

We measure activation dimensionality via PCA on the residual stream between blocks. On GPT-2, 90\% of residual stream variance after block 3 is captured by just 2 of 768 dimensions. However, projecting to these dimensions is catastrophic:

\begin{table}[h]
\centering
\caption{Effect of PCA projection on GPT-2 perplexity (blocks 3-8)}
\begin{tabular}{@{}lcc@{}}
\toprule
Dimensions kept & \% Variance & Perplexity \\
\midrule
8 & $\sim$95\% & 3,441 \\
64 & $\sim$99\% & 524 \\
256 & $\sim$99.9\% & 145 \\
768 (none) & 100\% & 47 (baseline) \\
\bottomrule
\end{tabular}
\end{table}

We verify this through CCA: the top-128 PCA directions and top-128 CCA directions (which maximize correlation between block 15 and block 31 outputs on Mistral) have only 4\% overlap. CCA directions achieve 23\% higher prediction $R^2$ than PCA directions at every dimensionality tested.

\begin{equation}
    \text{importance}(d) = \underbrace{\text{variance}(d)}_{\text{PCA captures}} \times \underbrace{\text{sensitivity}(d)}_{\text{PCA ignores}}
\end{equation}

Perturbation analysis confirms that sensitivity is approximately uniform across all 768 dimensions when averaged across tokens, while variance is highly concentrated. The product (importance) is therefore distributed broadly, with no safe directions to discard.

\begin{figure}[h]
\centering
\includegraphics[width=\textwidth]{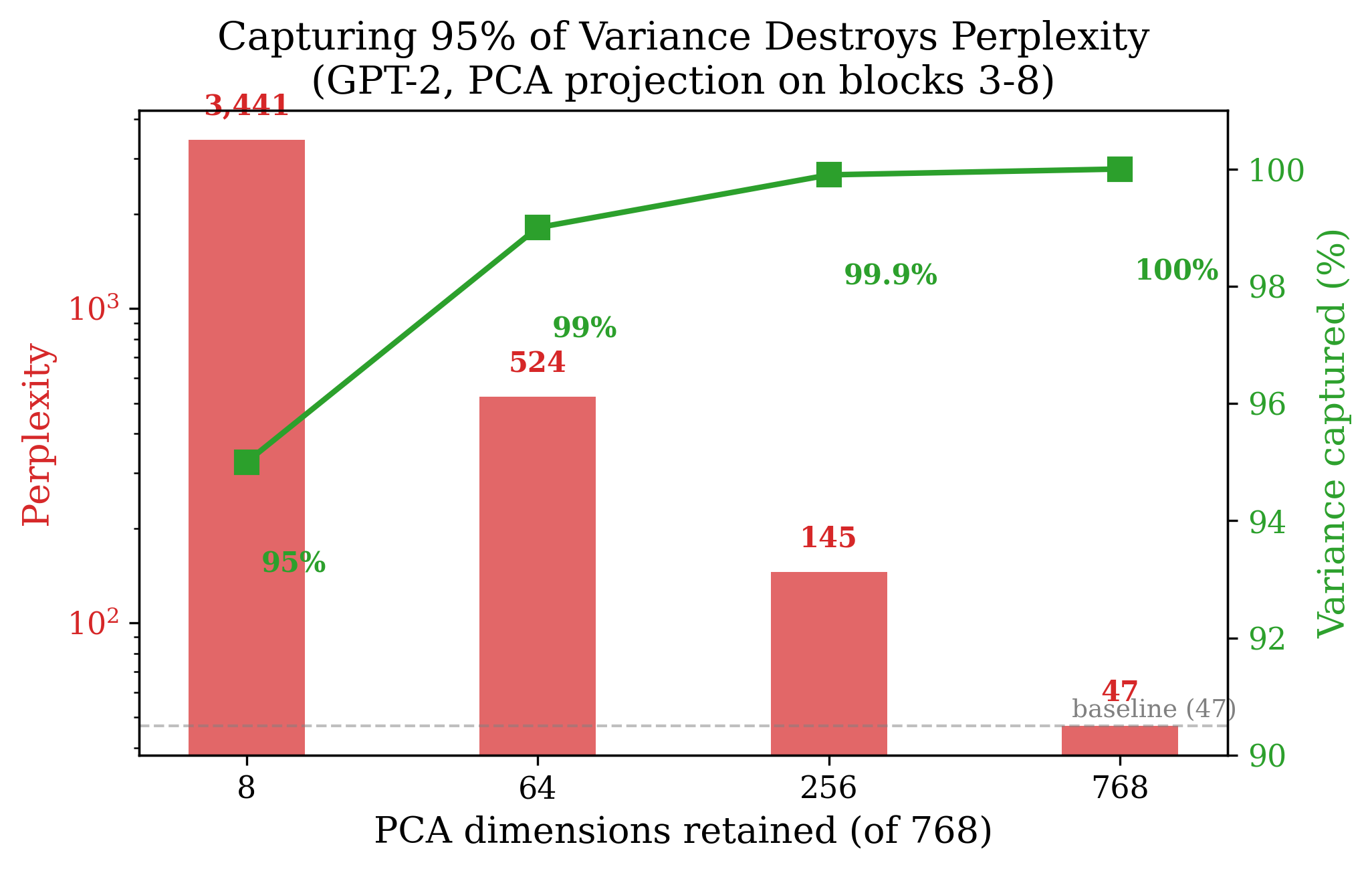}
\caption{Variance does not predict importance. Keeping 95\% of variance (8 PCA dimensions) yields catastrophic perplexity (3,441), while the baseline (all 768 dimensions) achieves 47. The gap demonstrates that high-variance directions are not the computationally important ones.}
\label{fig:variance_vs_ppl}
\end{figure}

\subsection{Finding 2: Block Linearity Is Conditional}

\begin{finding}
Transformer blocks are approximately linear on calibration data, but this linearity is conditional on the upstream activation distribution. Replacing upstream blocks shifts the distribution, degrading the linear approximation for downstream blocks.
\end{finding}

\begin{table}[h]
\centering
\caption{Block linearity ($R^2$) comparison across models}
\begin{tabular}{@{}lcc@{}}
\toprule
Block & GPT-2 $R^2$ & Mistral 7B $R^2$ \\
\midrule
0 (first) & $\approx 0.95$ & 0.168 \\
5 & $\approx 0.95$ & 0.452 \\
15 & $\approx 0.95$ & 0.580 \\
25 & $\approx 0.95$ & 0.638 \\
31 (last) & $\approx 0.95$ & 0.930 \\
\bottomrule
\end{tabular}
\end{table}

\begin{figure}[h]
\centering
\includegraphics[width=\textwidth]{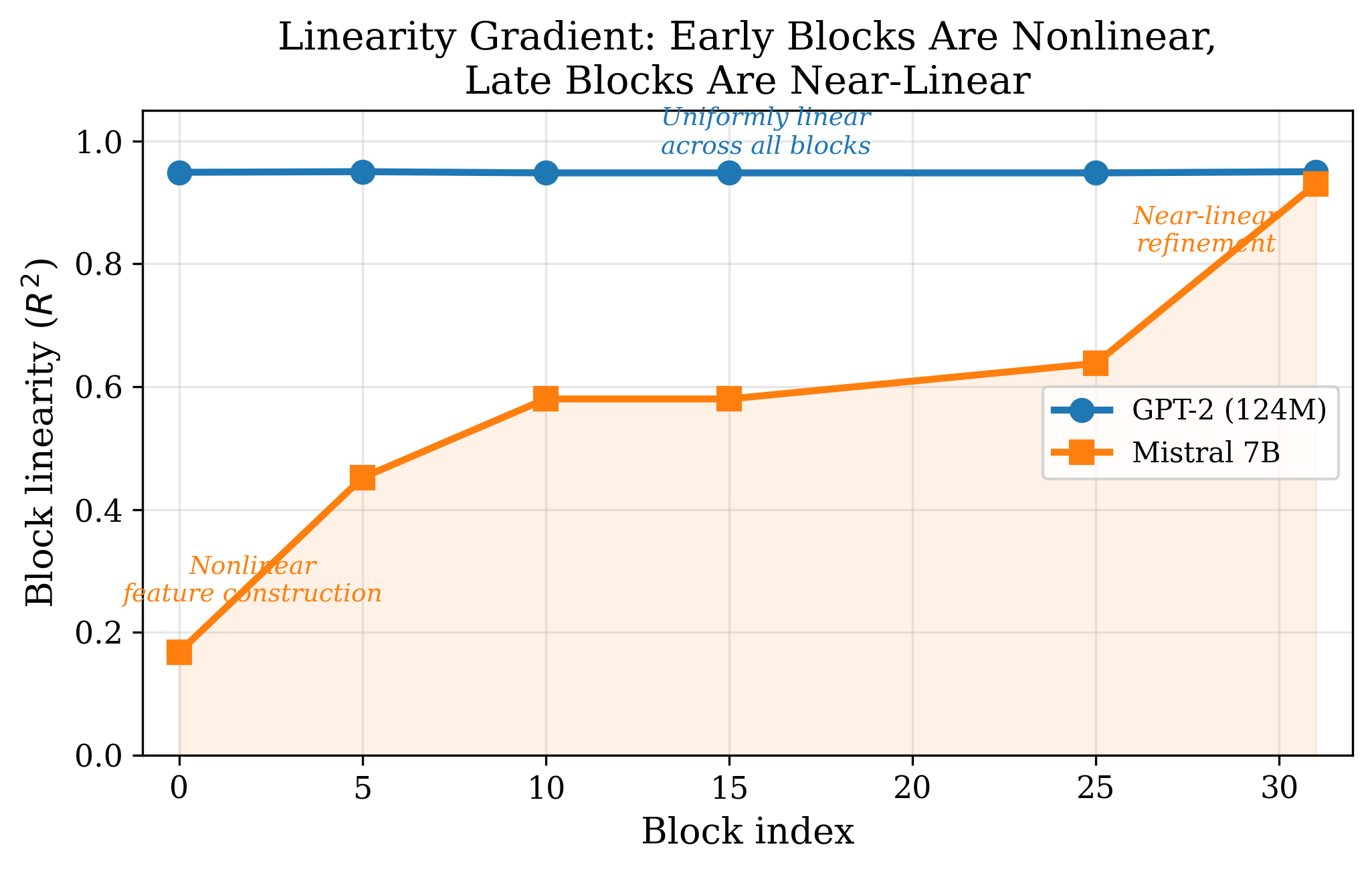}
\caption{Block linearity ($R^2$) across depth for GPT-2 and Mistral 7B. GPT-2 maintains uniformly high linearity (block-wise $R^2$ clustered around 0.95) at all depths. Mistral 7B shows a dramatic gradient from 0.17 (block 0) to 0.93 (block 31), revealing a division of labor between nonlinear feature construction and linear refinement.}
\label{fig:linearity_gradient}
\end{figure}

Single-block replacement works: on Mistral 7B, replacing block 31 with a rank-773 linear map (6.3M parameters replacing 218M) yields only +1.71 perplexity on held-out data. However, sequential multi-block replacement fails: $R^2$ degrades from 0.94 to 0.60 across 5 replaced blocks, with perplexity increasing by +151 after 5 blocks.

The degradation mechanism is twofold: (1) errors accumulate additively through residual connections ($x_{n+1} = x_n + f_n(x_n) + \epsilon_n$), and (2) the activation distribution shifts after each replacement, making subsequent linear fits less accurate.

\begin{figure}[h]
\centering
\includegraphics[width=\textwidth]{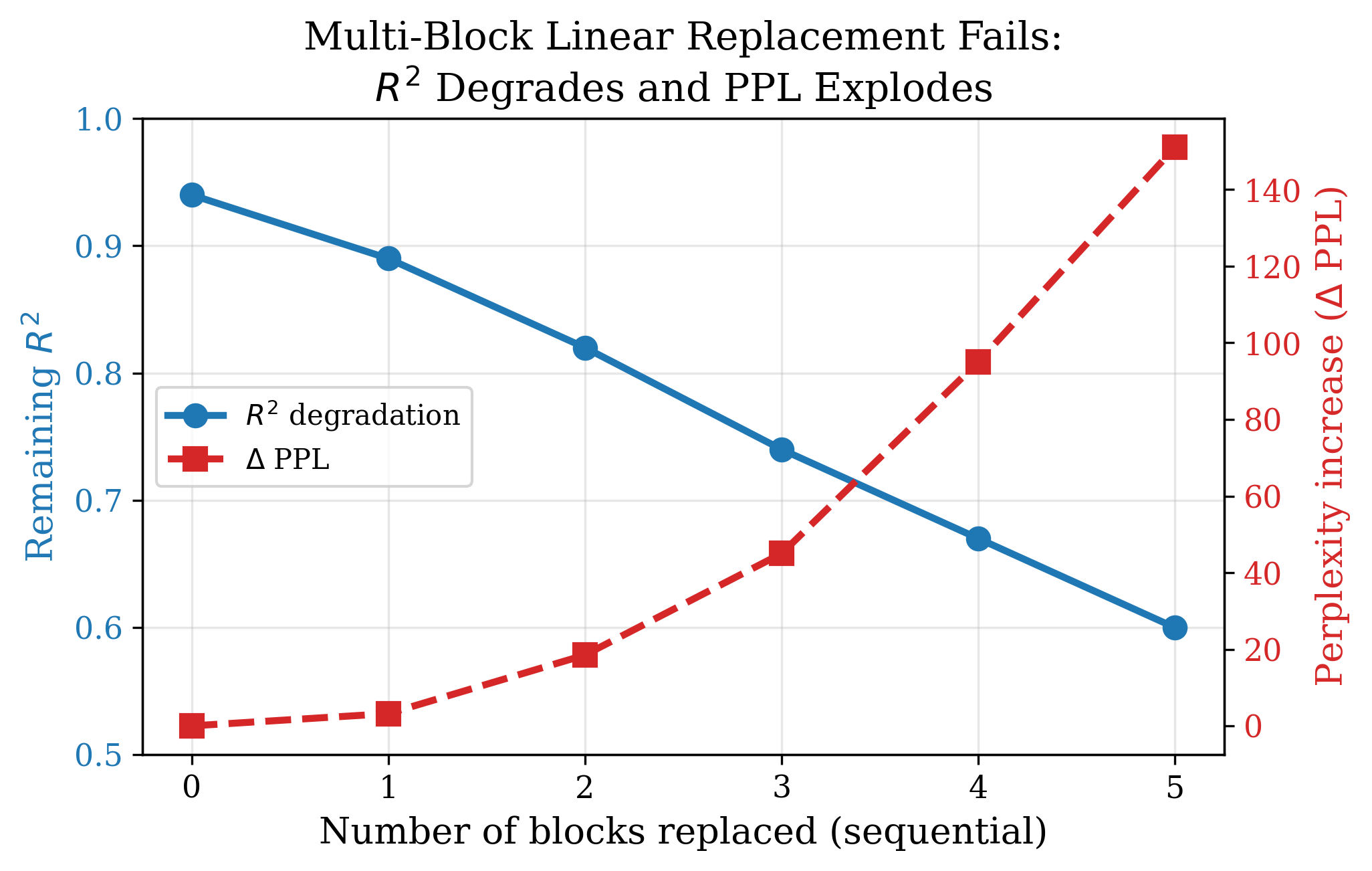}
\caption{Multi-block linear replacement fails catastrophically. $R^2$ degrades from 0.94 to 0.60 across 5 replaced blocks, while perplexity increases by +151. Each replacement shifts the activation distribution, degrading the linear approximation for downstream blocks.}
\label{fig:multiblock_failure}
\end{figure}

\subsection{Finding 3: The Reconstruction Wall}

\begin{finding}
Any approach that decomposes weights into quantized factors amplifies errors through cross-terms, making direct quantization strictly superior to factored approaches at the same bit budget.
\end{finding}

When weights are factored as $W = A \cdot B$ and each factor is independently quantized:
\begin{equation}
    \hat{W} = (A + \epsilon_A)(B + \epsilon_B) = AB + \epsilon_A B + A \epsilon_B + \epsilon_A \epsilon_B
\end{equation}

The cross-terms $\epsilon_A B$ and $A \epsilon_B$ typically exceed the error from directly quantizing $W$. We verify this empirically:

\begin{table}[h]
\centering
\caption{Factored vs direct quantization on GPT-2 (same 4-bit budget)}
\begin{tabular}{@{}lcc@{}}
\toprule
Method & Output MSE & vs INT4 \\
\midrule
INT4 (direct) & 0.000679 & baseline \\
DCT + INT4 & 0.001935 & 2.8$\times$ worse \\
SVD rank-384 + INT4 & 0.001359 & 2.0$\times$ worse \\
Rotated INT8/INT2 & 0.016882 & 24.8$\times$ worse \\
\bottomrule
\end{tabular}
\end{table}

Every factored approach is strictly worse than direct INT4 at the same 4-bit budget. The gap ranges from 2$\times$ (SVD+INT4) to 24.8$\times$ (rotated mixed-precision), confirming that the cross-term amplification in Eq.~4 is not a theoretical curiosity but the dominant source of error in practice.

\begin{figure}[h]
\centering
\includegraphics[width=\textwidth]{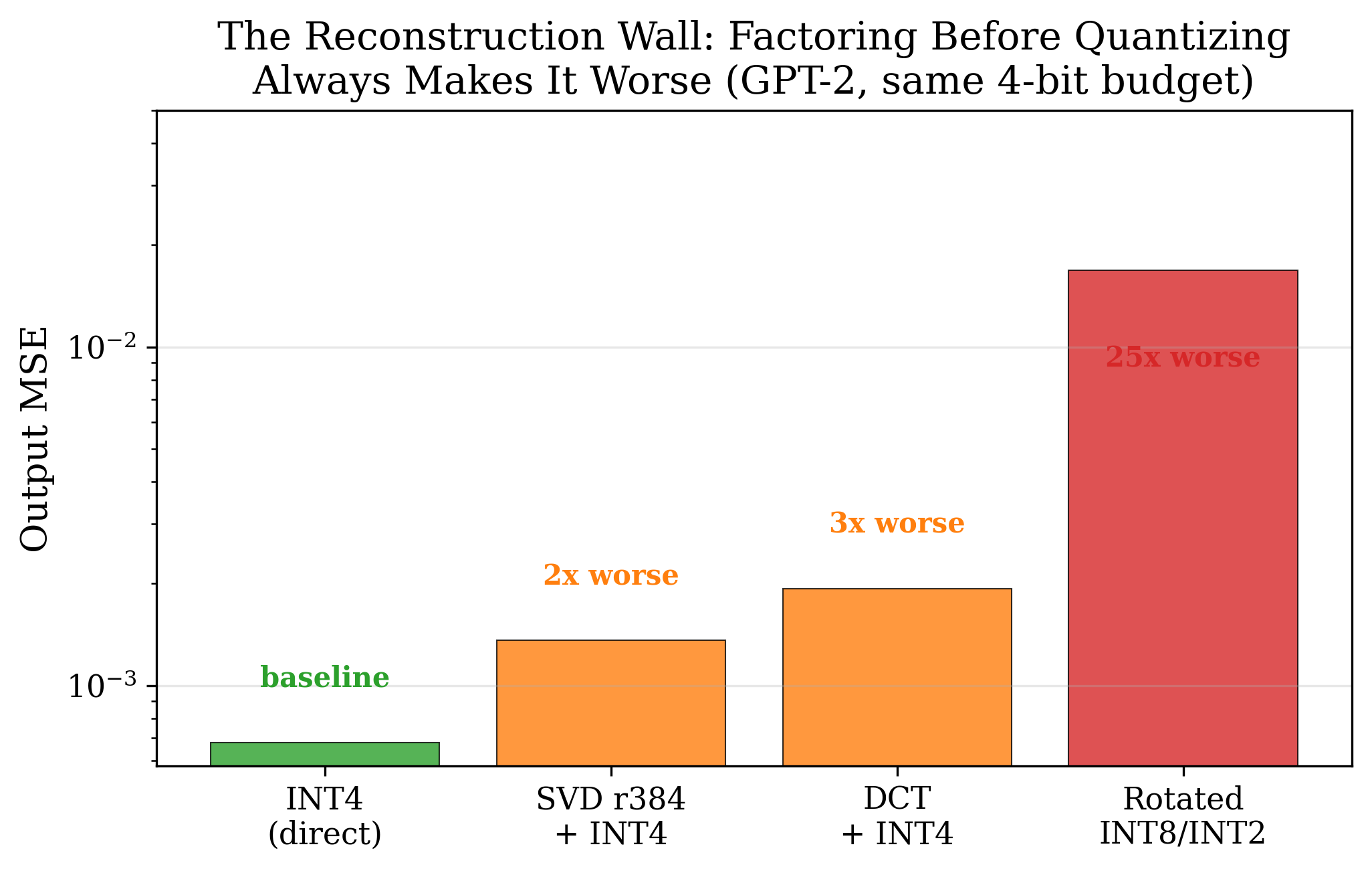}
\caption{The reconstruction wall: factored quantization amplifies errors through cross-terms. Direct INT4 achieves the lowest MSE; all factored approaches (DCT+INT4, SVD+INT4, Rotated INT8/INT2) are 2--25$\times$ worse at the same bit budget.}
\label{fig:reconstruction_wall}
\end{figure}

Note: GPTQ avoids this wall through sequential error feedback, which compensates for rounding errors during the quantization process. Our finding applies specifically to naive factored reconstruction without error compensation.

\subsection{Finding 4: Linearity Gradient Across Depth}

\begin{finding}
Mistral 7B exhibits a monotonic increase in block linearity with depth ($R^2$: 0.17 $\to$ 0.93), revealing a division of labor: early blocks perform nonlinear feature construction while late blocks perform near-linear refinement.
\end{finding}

This finding is consistent with the ``unreasonable ineffectiveness'' of deeper layers \cite{unreasonable} but adds a quantitative characterization via the $R^2$ metric.

Component-level analysis (KL divergence upon INT2 destruction) reveals that the importance is U-shaped, not monotonically decreasing:
\begin{itemize}
    \item Blocks 0-2: Critical (KL 3--4.5), dominated by attention
    \item Blocks 3-6: Important (KL 0.5--1.0), dominated by MLP
    \item Blocks 7-10: Cheapest (KL 0.23--0.27), both components light
    \item Block 11: Spike (KL 1.1), dominated by attention (prediction assembly)
\end{itemize}

\begin{figure}[h]
\centering
\includegraphics[width=\textwidth]{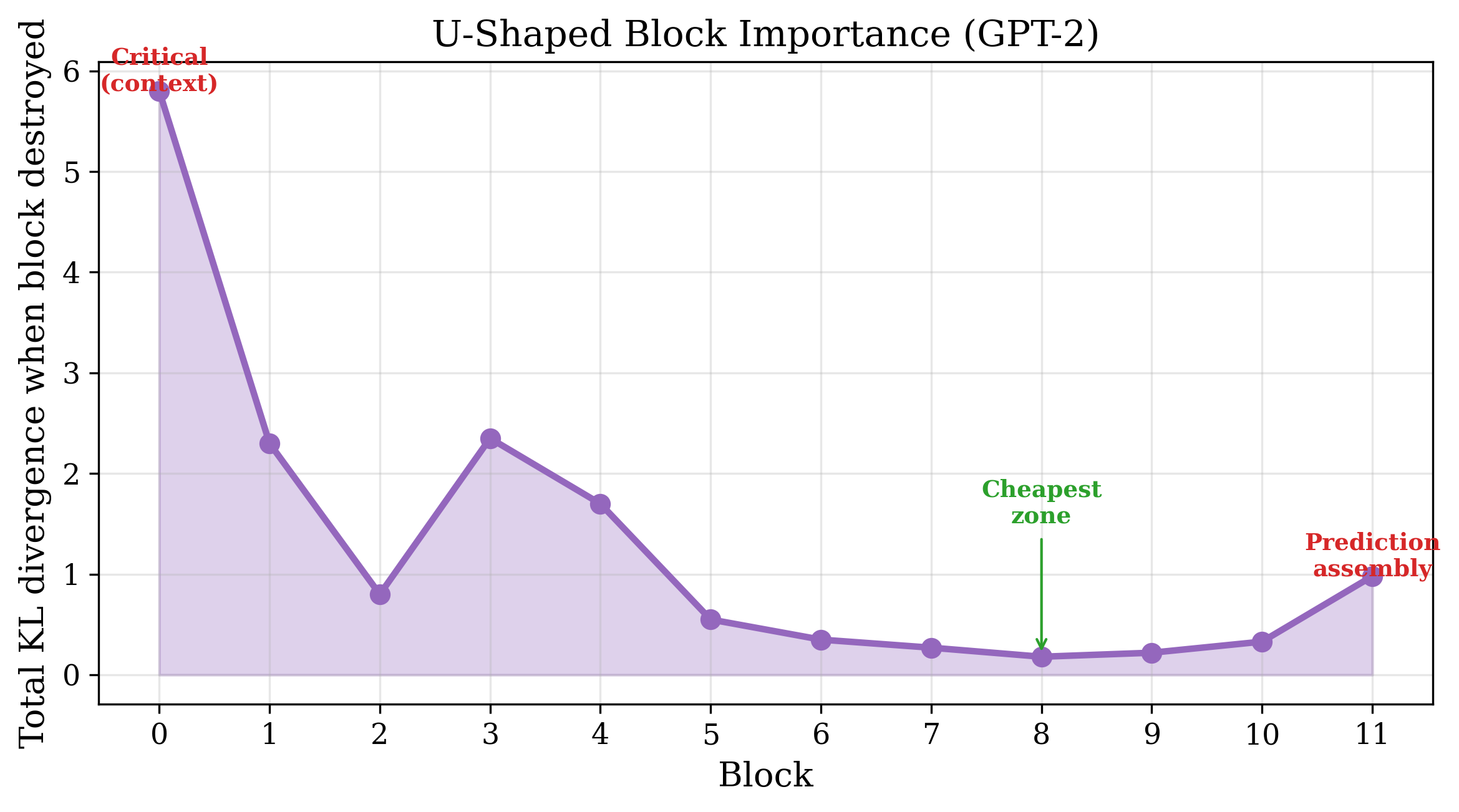}
\caption{Block-level importance (KL divergence upon INT2 destruction) reveals a U-shaped profile with four phases: context building (block 0), feature construction (blocks 2--4), refinement (blocks 5--9), and prediction assembly (block 11).}
\label{fig:block_importance}
\end{figure}

\subsection{Finding 5: 30\% of Tokens Are Easy}

\begin{finding}
Approximately 30\% of tokens can be classified as ``computationally easy,'' independently confirmed through two methods on two models.
\end{finding}

\textbf{Method 1: Trained exit heads.} On Mistral 7B, a dedicated prediction head trained at block 15 (500 steps, RMSNorm + Linear) achieves 30.2\% top-1 agreement with the full model. Extended training plateaus at $\sim$30\%.

\textbf{Method 2: KL sensitivity.} On GPT-2, 30.1\% of tokens show low KL divergence (below median) when late blocks (8--11) are destroyed via INT2 quantization.

Multi-exit routing with trained heads at blocks 15, 23, and 27 on Mistral yields:

\begin{table}[h]
\centering
\caption{Multi-exit routing: quality vs compute savings on Mistral 7B}
\begin{tabular}{@{}lccc@{}}
\toprule
Confidence threshold & Perplexity & $\Delta$ PPL & Compute saved \\
\midrule
1.0 (baseline) & 11.32 & --- & 0\% \\
0.95 & 15.63 & +4.31 & 6.4\% \\
0.80 & 21.98 & +10.66 & 13.3\% \\
0.70 & 27.35 & +16.04 & 17.4\% \\
0.50 & 47.07 & +35.75 & 26.9\% \\
\bottomrule
\end{tabular}
\end{table}

The quality-speed tradeoff is poor with post-hoc trained heads: even the safest threshold (0.95) incurs +4.31 PPL for only 6.4\% compute savings. The exit heads are confident about incorrect predictions. LoRA fine-tuning to push prediction information into early blocks (500 steps, 32 samples) yielded only +0.2\% improvement in this severely data-limited setting. Scaling to 10K steps on 866K tokens across three models (Section~\ref{sec:early_exit}) raises agreement to 35--53\%, suggesting that the 30\% easy-token ceiling reflects a training data limitation, not an architectural one.

\begin{figure}[h]
\centering
\includegraphics[width=\textwidth]{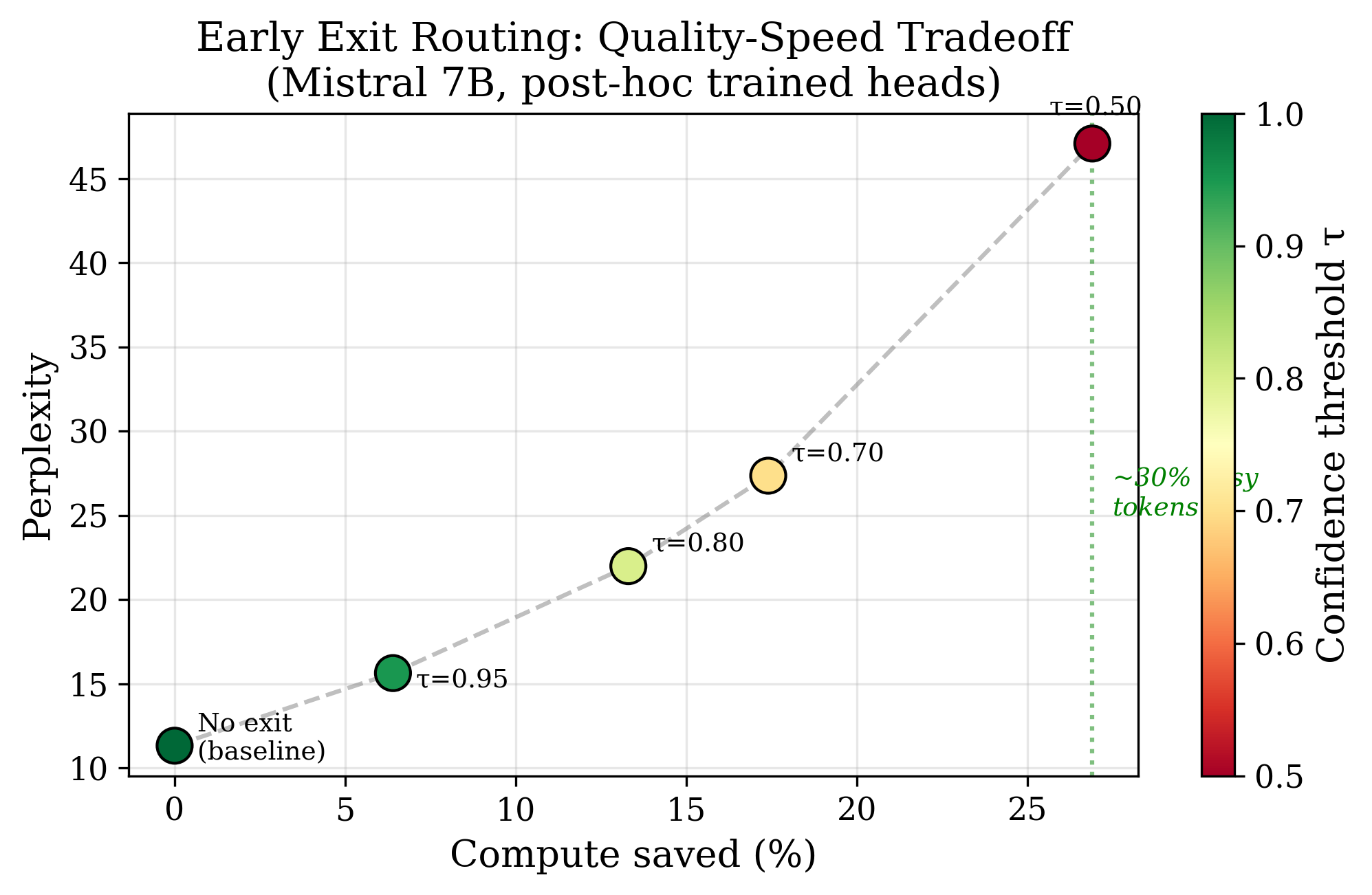}
\caption{Multi-exit routing tradeoff on Mistral 7B. The quality-compute curve shows diminishing returns: even conservative thresholds (0.95) incur significant perplexity cost (+4.31) for modest compute savings (6.4\%). The 30\% easy-token finding quantifies the ceiling for early-exit approaches.}
\label{fig:easy_tokens_routing}
\end{figure}

\section{Extended Analysis}

\subsection{Spectral Compression via DCT}

We applied the Discrete Cosine Transform (DCT) to weight matrices, hypothesizing that neural network weights—like images—might have concentrated frequency spectra amenable to JPEG-like compression.

\begin{definition}[Spectral Energy Concentration]
The Gini coefficient of the squared DCT coefficients measures how concentrated the spectral energy is. A Gini score near 1.0 indicates high concentration (compressible); near 0.0 indicates uniform spread (incompressible).
\end{definition}

\begin{table}[h]
\centering
\caption{DCT spectral energy concentration (Gini coefficient)}
\begin{tabular}{@{}lcc@{}}
\toprule
Layer type & Gini & Verdict \\
\midrule
Embeddings (wte, wpe) & 0.79--0.81 & Compressible \\
Transformer layers & 0.63--0.65 & Insufficient \\
Random matrix & $\sim$0.30 & Flat (control) \\
\bottomrule
\end{tabular}
\end{table}

\begin{figure}[h]
\centering
\includegraphics[width=\textwidth]{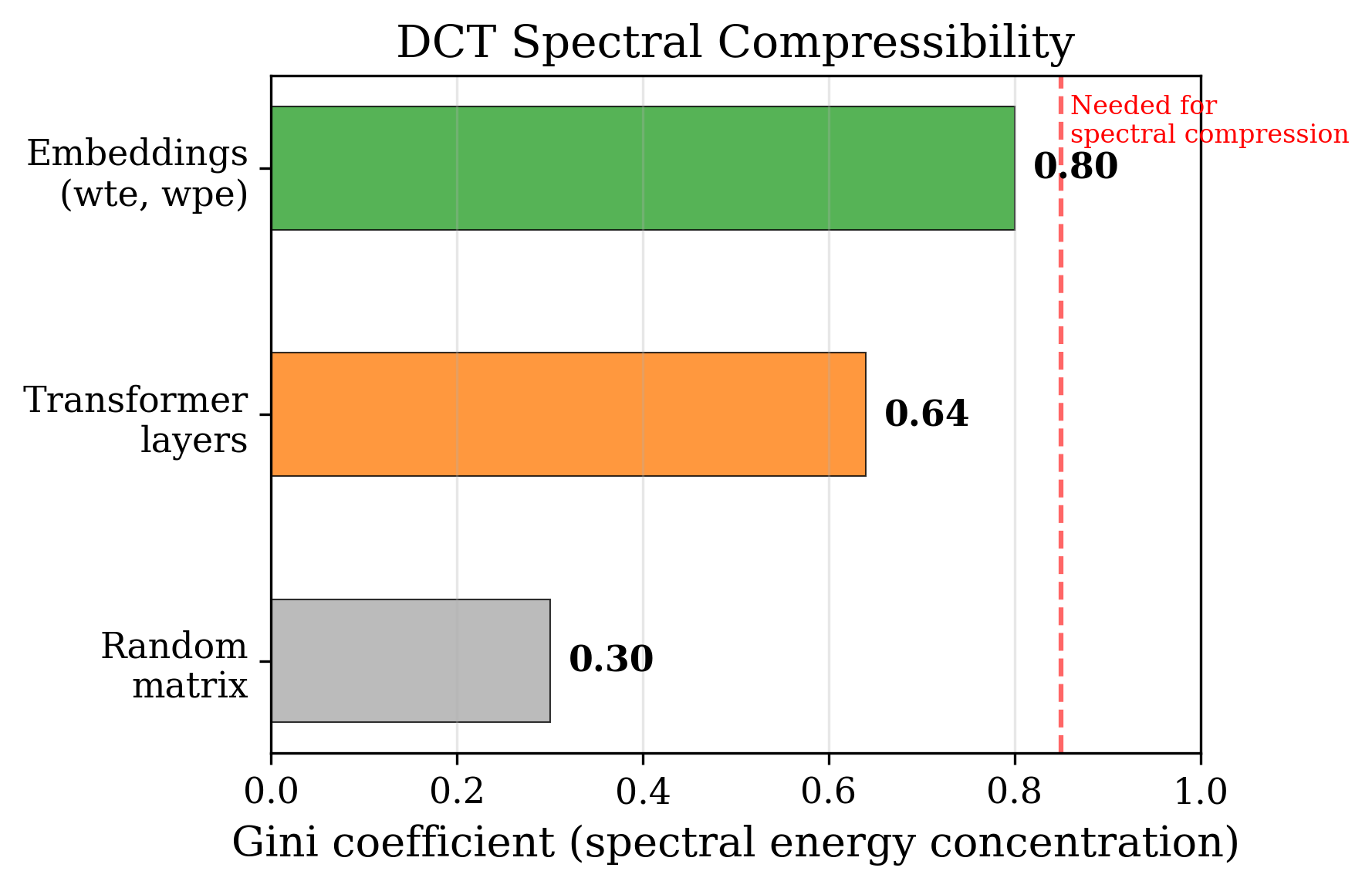}
\caption{DCT spectral energy concentration (Gini coefficient) across layer types. Embeddings show high concentration (0.79--0.81), transformer layers are moderate (0.63--0.65), and random matrices are flat (0.30). Transformer weights lack the spectral structure needed for effective frequency-domain compression.}
\label{fig:dct_gini}
\end{figure}

Transformer weight matrices show Gini scores of 0.63--0.65: above random (0.30) but insufficient for spectral compression to outperform direct quantization. The top 25\% of DCT coefficients capture only 72\% of energy, compared to $>$90\% needed for effective spectral compression. Embeddings show higher concentration (Gini 0.79--0.81) and are better candidates.

\subsection{K-means Quantization and NF4}

Standard INT4 quantization places 16 levels uniformly across $[w_{\min}, w_{\max}]$. K-means quantization places levels at the actual weight distribution peaks, which is provably optimal for minimum MSE (Lloyd-Max, 1957):

\begin{equation}
    \min_{c_1, \ldots, c_{16}} \sum_i \|w_i - c_{\text{nearest}(i)}\|^2
\end{equation}

\begin{table}[h]
\centering
\caption{K-means vs uniform INT4 on Mistral 7B (5 representative layers)}
\begin{tabular}{@{}lccc@{}}
\toprule
Layer & INT4 MSE & K-means MSE & Improvement \\
\midrule
layers.10.q\_proj & $1.7 \times 10^{-7}$ & $1.0 \times 10^{-7}$ & 1.6$\times$ \\
layers.20.gate\_proj & $1.2 \times 10^{-7}$ & $0.7 \times 10^{-7}$ & 1.6$\times$ \\
layers.25.up\_proj & $0.9 \times 10^{-7}$ & $0.6 \times 10^{-7}$ & 1.6$\times$ \\
layers.31.o\_proj & $1.4 \times 10^{-7}$ & $0.9 \times 10^{-7}$ & 1.6$\times$ \\
\bottomrule
\end{tabular}
\end{table}

K-means achieves a consistent 1.6$\times$ MSE improvement at the same 4-bit budget. However, this insight is not novel: NF4 (Normal Float 4) \cite{nf4}, used in QLoRA, pre-computes Gaussian-optimal quantization levels and achieves 1.21$\times$ better MSE at zero storage overhead. Approximately 60\% of k-means' advantage comes from Gaussian-optimal levels (captured by NF4) and 40\% from per-group adaptation (which incurs codebook storage overhead).

\subsection{Attention vs MLP: A Four-Phase Architecture}

By selectively destroying (INT2 quantization) the attention or MLP component of each block independently, we map the division of labor across depth. The resulting pattern reveals a clear four-phase functional architecture:

\begin{table}[h]
\centering
\caption{Component-level KL divergence when destroyed (GPT-2)}
\begin{tabular}{@{}lccl@{}}
\toprule
Depth & Attention KL & MLP KL & Dominant \\
\midrule
Block 0 & 4.7 & 1.1 & Attention (4.3$\times$) \\
Blocks 2--4 & 0.2--0.3 & 0.3--2.1 & MLP \\
Blocks 5--9 & 0.09--0.26 & 0.09--0.29 & Neither \\
Block 11 & 0.80 & 0.18 & Attention (4.5$\times$) \\
\bottomrule
\end{tabular}
\end{table}

\begin{figure}[h]
\centering
\includegraphics[width=\textwidth]{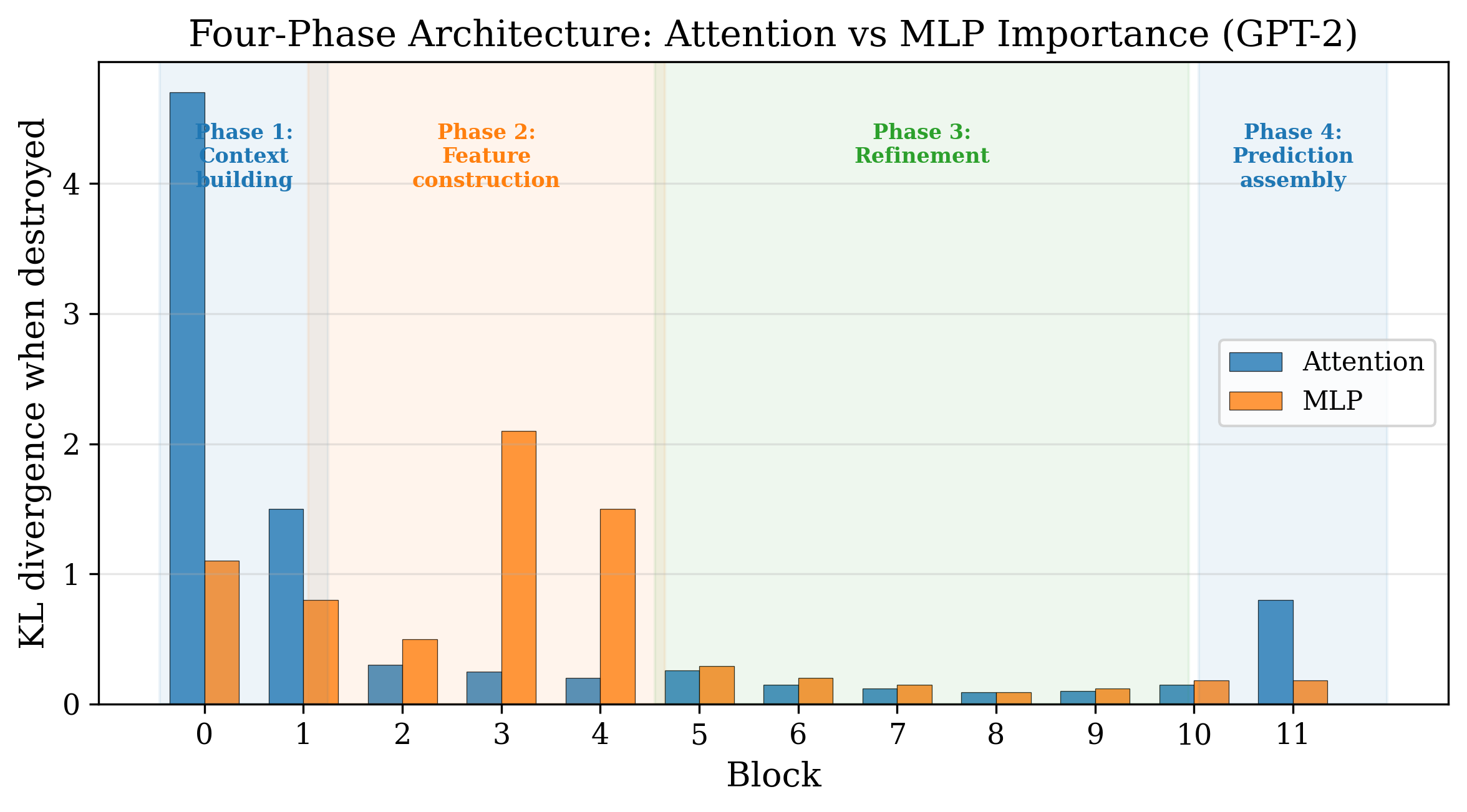}
\caption{Four-phase functional architecture of GPT-2, revealed by component-level destruction (INT2 quantization). Phase I: context building (attention-dominated). Phase II: feature construction (MLP-dominated). Phase III: refinement (both light). Phase IV: prediction assembly (attention-dominated).}
\label{fig:attention_mlp_phases}
\end{figure}

The model exhibits a clear four-phase architecture:
\begin{enumerate}
    \item \textbf{Context building} (block 0): Attention-dominated. Establishes token relationships.
    \item \textbf{Feature construction} (blocks 2--4): MLP-dominated. Builds representations.
    \item \textbf{Refinement} (blocks 5--9): Both components light. Gentle polishing.
    \item \textbf{Prediction assembly} (block 11): Attention-dominated. Assembles final predictions.
\end{enumerate}

Notably, block 11's MLP contributes KL=0.18 (low importance) while its attention contributes KL=0.80. However, skipping block 11's MLP entirely still costs +3.05 perplexity on Mistral (block 31), and replacing it with its cached mean output performs even worse (+4.51), indicating that the MLP's value lies in its per-token variation, not its average output.

\subsection{Early Exit: From Observation to Routing}
\label{sec:early_exit}

We train dedicated prediction heads (RMSNorm + Linear, initialized from the final model head) at blocks 15, 23, and 27 of Mistral 7B. Each head is trained for 500 steps on 8K calibration tokens.

\begin{table}[h]
\centering
\caption{Early exit head agreement with full model (Mistral 7B)}
\begin{tabular}{@{}lccc@{}}
\toprule
Exit point & Naive head & Trained head & Blocks saved \\
\midrule
Block 15 & 5.5\% & 30.2\% & 53\% of compute \\
Block 23 & 19.6\% & 40.4\% & 28\% of compute \\
Block 27 & 33.7\% & 49.4\% & 16\% of compute \\
\bottomrule
\end{tabular}
\end{table}

\begin{figure}[h]
\centering
\includegraphics[width=\textwidth]{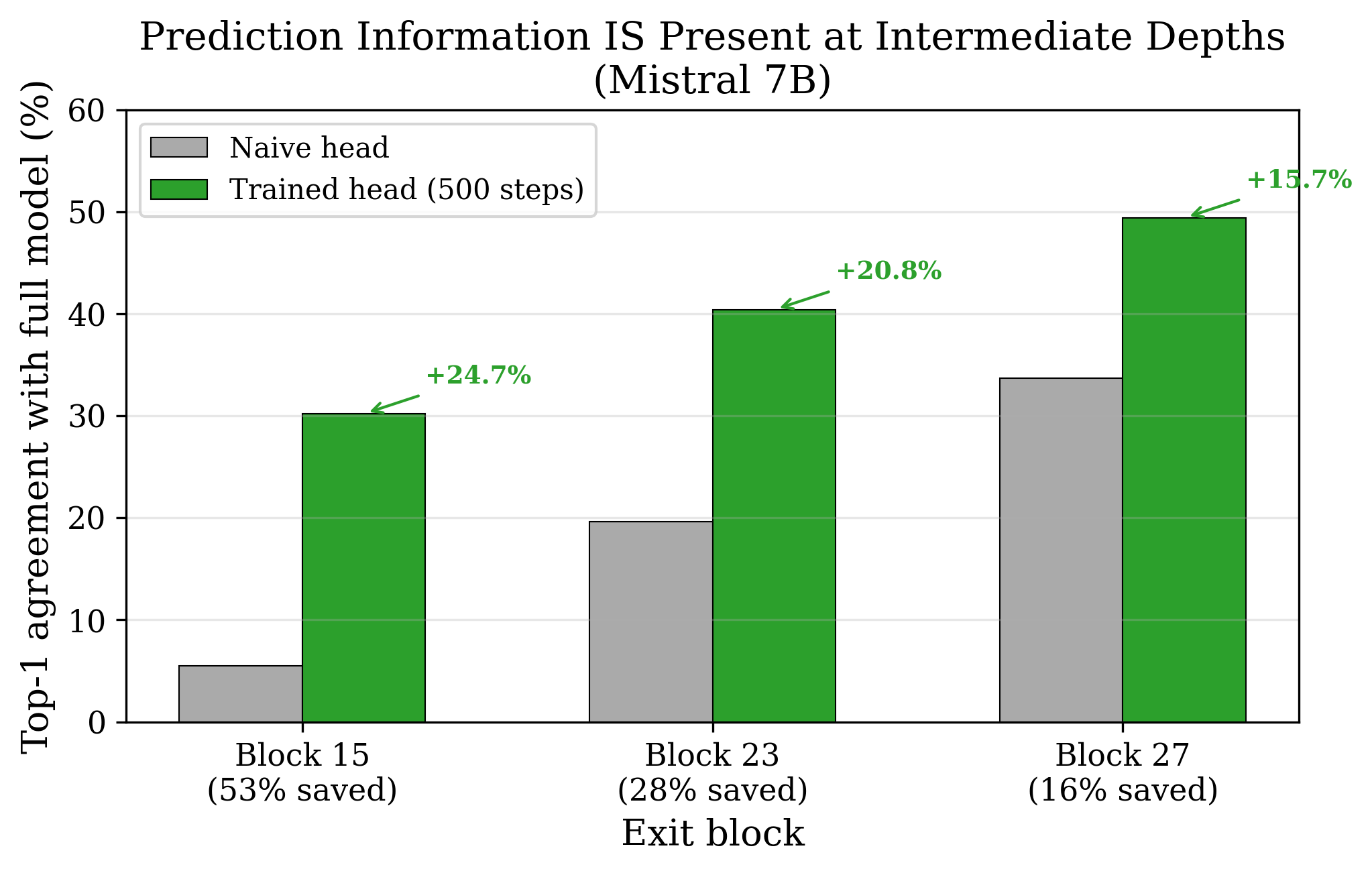}
\caption{Early exit head agreement with full model on Mistral 7B. Naive heads (untrained) achieve poor agreement, while trained heads (500 steps) show substantial improvement---demonstrating that prediction-relevant information is present at intermediate depths but encoded differently than the final head expects.}
\label{fig:exit_heads}
\end{figure}

The naive-to-trained improvement (5.5\% $\to$ 30.2\% at block 15) demonstrates that prediction-relevant information \textit{is present} at intermediate depths but encoded differently than the final head expects.

Multi-exit confidence-based routing (exit at the earliest block where softmax confidence exceeds a threshold) yields compute savings of 6.4--26.9\%, but at significant perplexity cost (+4.31 to +35.75). The quality-speed tradeoff is poor with post-hoc trained heads, as high softmax confidence does not reliably indicate prediction correctness on held-out data.

LoRA fine-tuning (rank 64, dual loss $\mathcal{L} = \mathcal{L}_{\text{final}} + 0.5 \cdot \mathcal{L}_{\text{early}}$, 500 steps on 32 samples) improved agreement by only +0.2\% in this data-limited setting. Scaling to proper fine-tuning conditions (10K steps, 866K tokens, three models including Mistral 7B, Llama 3.2 3B, and TinyLlama 1.1B) raises agreement to 35--53\%. Cross-entropy supervision consistently outperforms KL distillation for exit head training, suggesting that top-token fidelity, not distribution matching, is the correct optimization target for early exit systems.

\subsection{Rotation-Based Quantization Fails Similarly}

We additionally test whether rotating into an activation-informed basis (PCA or CCA) followed by mixed-precision quantization (INT8 on important rows, INT2 on unimportant rows) can circumvent the reconstruction wall. Despite CCA providing 23\% better prediction $R^2$ than PCA (Section 4.1), both rotated mixed-precision schemes produce 20--21$\times$ worse MSE than direct INT4 at the same 4-bit budget. The reconstruction wall applies regardless of basis choice. This is consistent with the basis ablation in \cite{salfati2026quant}, which shows $<$0.4 PPL spread across four different bases under direct quantization: the advantage of quantization over rank reduction comes from preserving all dimensions, not from choosing a better coordinate system.

\section{Practical Implications}

We summarize the actionable guidance that follows from each structural finding:

\begin{table}[h]
\centering
\caption{From structural findings to practitioner guidance}
\begin{tabular}{@{}p{3.8cm}p{5.5cm}p{4.5cm}@{}}
\toprule
\textbf{Finding} & \textbf{Implication} & \textbf{Recommendation} \\
\midrule
1. Variance $\neq$ importance & PCA, activation magnitude, and output variance are poor proxies for which dimensions to compress & Use downstream-aware importance metrics or compress uniformly via quantization \\
\addlinespace
2. Conditional linearity & Single-block linear replacement works; multi-block fails due to distribution shift & Limit linear approximation to $\leq$1 contiguous block; prefer direct quantization for multi-block compression \\
\addlinespace
3. Reconstruction wall & Factored quantization amplifies errors through cross-terms & Quantize weights directly (GPTQ, AWQ) rather than factoring first \\
\addlinespace
4. Linearity gradient & Early blocks are nonlinear (must preserve); late blocks are near-linear (safe to compress) & Allocate compression budget unevenly: protect early blocks, compress late blocks more aggressively \\
\addlinespace
5. 30\% easy tokens & A substantial fraction of tokens require minimal computation & Invest in adaptive per-token computation (early exit, speculative decoding) rather than static compression alone \\
\bottomrule
\end{tabular}
\end{table}

The overarching message is that \textit{compression should respect the geometry of the computation, not the geometry of the representation}. Methods that operate uniformly on all directions (quantization) outperform methods that selectively discard directions (rank reduction), because the directions that appear dispensable by energy-based criteria are often computationally critical.

\begin{figure}[h]
\centering
\includegraphics[width=\textwidth]{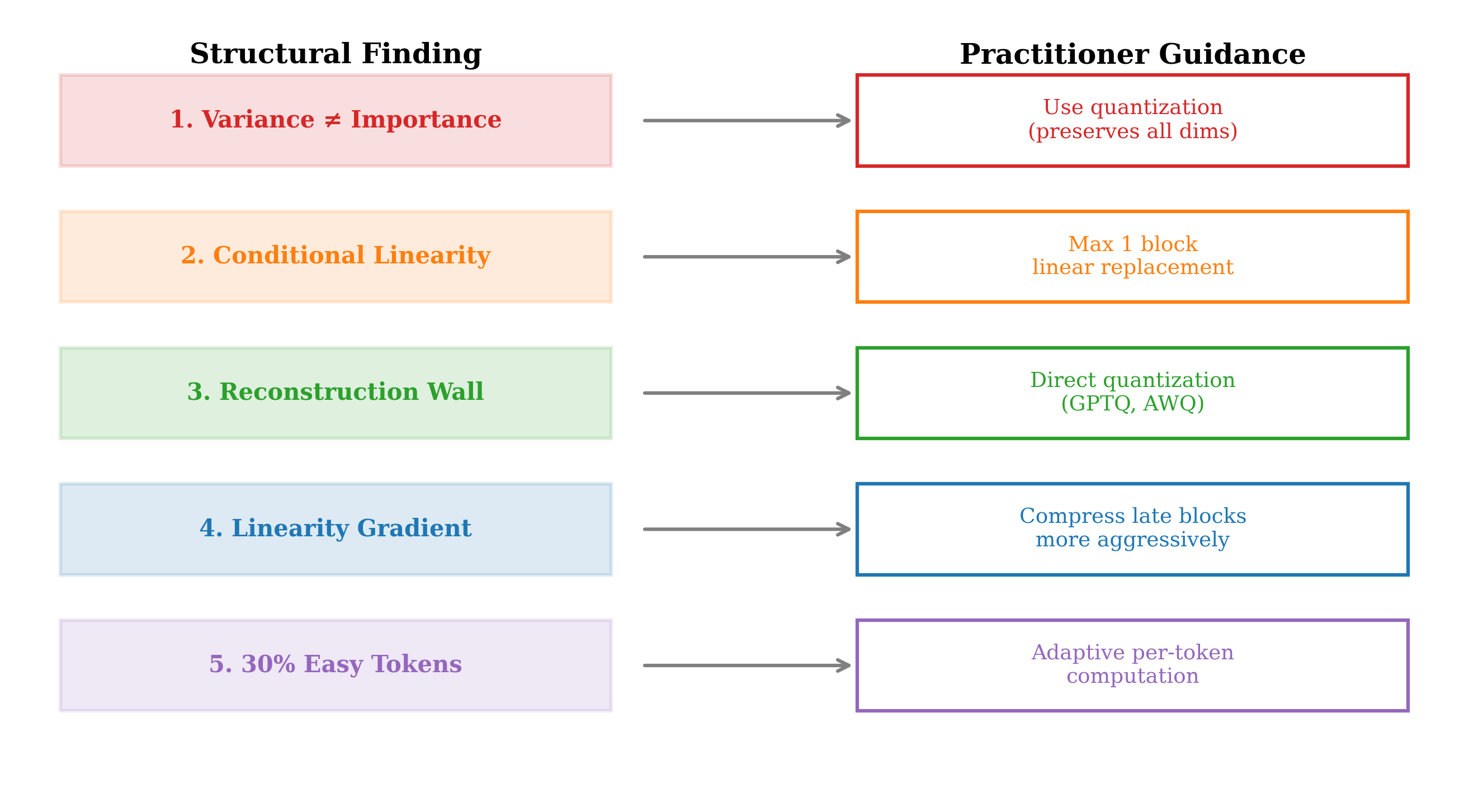}
\caption{Summary of five structural findings and their implications for transformer compression. Each finding maps to a specific failure mode of existing compression methods and suggests an alternative approach.}
\label{fig:summary}
\end{figure}

\section{Discussion}

\subsection{Why Static Compression Is Hard}

Our experiments reveal a fundamental tension: the structural properties that make individual blocks compressible (approximate linearity, concentrated variance) do not compose across blocks. The residual connection architecture preserves all errors additively, while each approximation shifts the activation distribution for downstream computation. This is not a failure of any particular compression method---it is a structural property of the residual stream itself.

The practical ceiling is set by direct quantization methods (GPTQ, AWQ), which sidestep the composition problem by operating uniformly on all directions within each layer. The ``free lunch'' of exploiting structural redundancy appears limited to single blocks (3\% of the model) or marginal quantization improvements (NF4's 1.2$\times$ better MSE). Our companion paper \cite{salfati2026quant} confirms this empirically across five models: quantization outperforms rank reduction by 4--364 PPL at matched storage budgets.

\subsection{The Promise of Adaptive Computation}

The 30\% easy-token finding suggests a different path: rather than compressing the model statically, allocate computation dynamically per token. Easy tokens (30\%) could exit early at intermediate layers, while hard tokens receive full computation. The linearity gradient provides a structural basis for this: later blocks (high $R^2$) perform near-linear refinement that may be unnecessary for tokens already confidently predicted.

However, our experiments show this requires more than post-hoc analysis---the model must be fine-tuned to support early exit, either through multi-exit training objectives or LoRA-based dual-loss approaches that push prediction-relevant information into early blocks.

\subsection{Variance vs Importance: Implications Beyond Compression}

The strong decorrelation between PCA and CCA directions (96\% different) has implications beyond compression. It suggests that analyses based on activation variance---a common tool in interpretability research, pruning, and feature attribution---may systematically misidentify which dimensions are functionally important. Any method that equates ``large activation'' with ``important activation'' is vulnerable to this failure mode.

\subsection{Limitations}

Our study has several limitations. (1) We test only two models (GPT-2 124M and Mistral 7B); while these span 58$\times$ in scale and differ in architecture (MHA vs GQA), additional architectures are needed. (2) All experiments use WikiText-2; cross-domain validation would strengthen the findings. (3) Evaluation uses 16 calibration samples for most experiments---sufficient for the structural measurements reported but below the $\geq$1,000-chunk threshold we recommend for perplexity evaluation in deployment contexts. (4) We do not test models larger than 7B, where compression behavior may differ.

\section{Conclusion}

Through over 40 systematic experiments across two model scales, we establish five structural properties of transformer compressibility. Our central finding is that \textit{variance is not importance}: the standard tools for identifying redundancy (PCA, activation magnitude, output variance) are poor proxies for computational importance. Direct quantization methods avoid this issue by operating uniformly on all directions, while more sophisticated approaches that attempt to exploit structural redundancy consistently fall into the reconstruction wall or composition failure.

We additionally identify a clear four-phase functional architecture (context building $\to$ feature construction $\to$ refinement $\to$ prediction assembly) that suggests compression strategies should be phase-aware rather than uniform across depth.

The most promising direction for inference cost reduction is not static compression but adaptive computation: dynamically allocating depth and precision per token based on prediction confidence. The 30\% easy-token finding quantifies the opportunity, while the linearity gradient ($R^2$: 0.17 $\to$ 0.93) identifies which blocks are safe to bypass. Realizing this potential requires integration with model training, as demonstrated by the scaled LoRA fine-tuning results in Section~\ref{sec:early_exit}.

\section*{Code and Data Availability}

All experiment scripts, calibration data splits, and raw perplexity / CCA / R$^2$ outputs underlying the tables and figures in this paper are available upon request from the corresponding author. A public release accompanying the companion paper \cite{salfati2026quant} is planned.

\bibliographystyle{plain}

\end{document}